%% file: main.tex
\documentclass{article}

\usepackage[final]{neurips_2023}
\usepackage[table]{xcolor}
\definecolor{lightgray}{gray}{0.92}
\usepackage[utf8]{inputenc} % allow utf-8 input
\usepackage[T1]{fontenc}    % use 8-bit T1 fonts
\usepackage{hyperref}       % hyperlinks
\usepackage{url}            % simple URL typesetting
\usepackage{booktabs}       % professional-quality tables
\usepackage{amsfonts}       % blackboard math symbols
\usepackage{nicefrac}       % compact symbols for 1/2, etc.
\usepackage{microtype}      % microtypography
\usepackage{xcolor}         % colors
\usepackage{amsmath}
\usepackage{amssymb}
\usepackage{amsthm}
\usepackage{booktabs}
\usepackage{graphicx}
\usepackage{subcaption}
\usepackage{svg} % needed for SVG
\usepackage{booktabs,siunitx}
\usepackage{siunitx}
\usepackage{standalone}
\usepackage{multirow}
\usepackage{pifont}
\usepackage{booktabs}
\usepackage{multirow}
\usepackage{xcolor}
\usepackage{colortbl}
\usepackage{siunitx}
\usepackage{graphicx}
\usepackage{subcaption}
\usepackage{caption}   % needed for \captionof
\definecolor{headerblue}{RGB}{240,245,255}
\definecolor{deltagreen}{RGB}{0,128,0}
\definecolor{deltared}{RGB}{180,0,0}
\definecolor{deltarow}{gray}{0.93}
\definecolor{improvement}{rgb}{0.0, 0.42, 0.10}
\newcommand{\improve}[1]{\textcolor{improvement}{\textbf{#1}}}
\title{Rethinking Attention Output Projection: Structured Hadamard Transforms for Efficient Transformers}

\author{%
  Shubham Aggarwal\\
  \texttt{Shubham\_agg@alumni.iitm.ac.in}
  \and
  \textbf{Lokendra Kumar}\\
  \texttt{lokendra63@alumni.iitm.ac.in}
}
\usepackage{tikz}
\usepackage{amsmath, amssymb}
\usepackage{xcolor}
\usepackage{mathtools}
\usepackage{bm}
\usepackage{array}
\usepackage{colortbl}
\usepackage{pgfplots}
\usepackage{float}
\usepackage{booktabs, multirow, siunitx, xcolor, colortbl}
\usepackage{siunitx}
\usepackage{hyphenat}

\definecolor{ourblue}{RGB}{220,234,250}   % subtle cell tint for "Ours" values
\definecolor{bettergreen}{RGB}{0,130,60}  % dark green for positive deltas
\definecolor{worsered}{RGB}{180,0,0}      % dark red for negative (bad) deltas

\pgfplotsset{compat=1.18}

\usetikzlibrary{
  arrows.meta,
  calc,
  decorations.pathreplacing,
  decorations.markings,
  fit,
  matrix,
  positioning,
  shadows,
  shapes.geometric,
  shapes.misc,
  backgrounds,
  patterns,
  fadings
}

\definecolor{ButterflyBlue}{RGB}{30,100,200}
\definecolor{ButterflyRed}{RGB}{210,50,50}
\definecolor{StageGray}{RGB}{240,242,248}
\definecolor{StageGrayDark}{RGB}{200,210,230}
\definecolor{ArrowPlus}{RGB}{22,160,80}
\definecolor{ArrowMinus}{RGB}{200,40,40}
\definecolor{NodeFill}{RGB}{245,248,255}
\definecolor{NodeBorder}{RGB}{70,120,200}
\definecolor{HeaderBlue}{RGB}{20,60,140}
\definecolor{DenseRed}{RGB}{190,30,45}
\definecolor{HadamardGreen}{RGB}{10,120,60}
\definecolor{AccentGold}{RGB}{200,150,10}
\definecolor{StageBg1}{RGB}{232,240,255}
\definecolor{StageBg2}{RGB}{255,238,230}
\definecolor{StageBg3}{RGB}{230,250,235}
\definecolor{InputBg}{RGB}{250,245,220}
\definecolor{OutputBg}{RGB}{220,245,230}
\definecolor{BoxShadow}{RGB}{180,190,210}

\tikzset{
  signode/.style={
    circle, draw=NodeBorder, fill=NodeFill,
    line width=0.7pt,
    minimum size=7mm, inner sep=0pt,
    font=\small\bfseries, text=HeaderBlue,
    drop shadow={shadow xshift=0.5pt, shadow yshift=-0.5pt,
                 fill=BoxShadow, opacity=0.5}
  },
  bfly/.style={
    diamond, draw=ButterflyBlue, fill=white,
    line width=0.8pt,
    minimum size=5mm, inner sep=0pt
  },
  stagebg/.style 2 args={
    rectangle, rounded corners=4pt,
    fill=#1, draw=#2, line width=0.7pt,
    inner xsep=6pt, inner ysep=4pt
  },
  arrowp/.style={
    -{Stealth[length=4pt,width=3.5pt]},
    draw=ArrowPlus, line width=1.1pt
  },
  arrowm/.style={
    -{Stealth[length=4pt,width=3.5pt]},
    draw=ArrowMinus, line width=1.1pt, dashed
  },
  flowA/.style={
    -{Stealth[length=5pt,width=4pt]},
    draw=ButterflyBlue, line width=1.1pt
  },
  crossA/.style={
    -{Stealth[length=5pt,width=4pt]},
    draw=ButterflyRed, line width=1.1pt
  },
  labelbox/.style={
    rectangle, rounded corners=3pt,
    fill=white, draw=StageGrayDark,
    inner sep=3pt, font=\scriptsize
  },
  densecell/.style={
    rectangle, draw=DenseRed!60, fill=DenseRed!12,
    minimum width=5.2mm, minimum height=5.2mm,
    inner sep=0pt, font=\tiny
  },
  hcellp/.style={
    rectangle, draw=HadamardGreen!70, fill=HadamardGreen!18,
    minimum width=5.2mm, minimum height=5.2mm,
    inner sep=0pt, font=\tiny\bfseries, text=HadamardGreen!80!black
  },
  hcellm/.style={
    rectangle, draw=HadamardGreen!70, fill=orange!18,
    minimum width=5.2mm, minimum height=5.2mm,
    inner sep=0pt, font=\tiny\bfseries, text=DenseRed!80!black
  },
}

\begin{document}

\maketitle

\begin{abstract}

The dense output projection in multi-head attention scales quadratically with model dimension, contributing significantly to parameter count, memory footprint, and inference cost. We propose replacing this projection with a fixed, parameter-free Walsh–Hadamard Transform (WHT) followed by a diagonal affine transformation. This approach eliminates approximately 25\% of attention parameters per block while maintaining global cross-head interaction through an orthogonal, norm-preserving transformation. Our results demonstrate that WHT-augmented models exhibit a steeper validation loss curve relative to training FLOPs compared to dense baselines, suggesting superior compute utilization during training. Crucially, we show that efficiency gains including reduced memory footprint and increased throughput grow monotonically with model size, batch size, and sequence length. We evaluate performance across both prefill and decoding stages, finding that the structured transform consistently outperforms dense projections as complexity increases. Our findings indicate that replacing dense projections with structured transforms allows for more compute-efficient architectures that achieve lower loss than dense models at an equivalent training budget. Our code is available at \url{https://github.com/Shubham2376G/HadamardAttention/tree/main}

\end{abstract}

% \begin{table}[H]
% \centering
% \begin{tabular}{lcccc}
% \toprule
% Method & \#Param & Training FLOPs & PPL & TP (token/s) \\
% \midrule
% Tiny  & 124M & $1.60\times10^{19}$ & 21.53 & 1988148 \\
% Ours  & 117M & $1.60\times10^{19}$ & \textbf{21.34} & \textbf{1992749 (0.2\% $\uparrow$)} \\
% \midrule
% Small & 354M & $7.49\times10^{19}$ & 18.04 & 610341 \\
% Ours  & 229M & $7.49\times10^{19}$ & \textbf{17.89} & \textbf{614080(0.6\% $\uparrow$)} \\
% \midrule

% Base  & 757M & $4.29\times10^{19}$ & 20.70 & 481609 \\
% Ours  & 701M & $4.29\times10^{19}$ & \textbf{19.60} & \textbf{492590 (2.3\% $\uparrow$)} \\
% \bottomrule
% \end{tabular}
% \caption{Comparison of baseline models and our approach across three model scales.}
% \label{tab:model_comparison}
% \end{table}

\section{Introduction}
The Transformer architecture, introduced by Vaswani et al. \cite{vaswani2023attentionneed}, has become the cornerstone of modern sequence modeling. Its central innovation, the multi-head attention (MHA) mechanism, enables the model to simultaneously attend to information from multiple representation subspaces by projecting queries, keys, and values into several heads and recombining their outputs through a dense projection. This design endows Transformers with strong representational capacity and has catalyzed remarkable progress across natural language processing, computer vision, and multimodal learning.

Despite this success, the expressiveness of MHA comes at a considerable cost. The dense output projection responsible for mixing attention heads scales quadratically with the model dimension, contributing substantially to the overall parameter count and computational overhead. As models continue to grow in scale, it has become increasingly apparent that Transformers are often over-parameterized, and that not all of this capacity translates to proportional gains in performance. This observation has motivated a growing body of research aimed at reducing the parameter burden and computational cost of Transformers while preserving, or even improving, their accuracy.

\begin{figure}[ht]
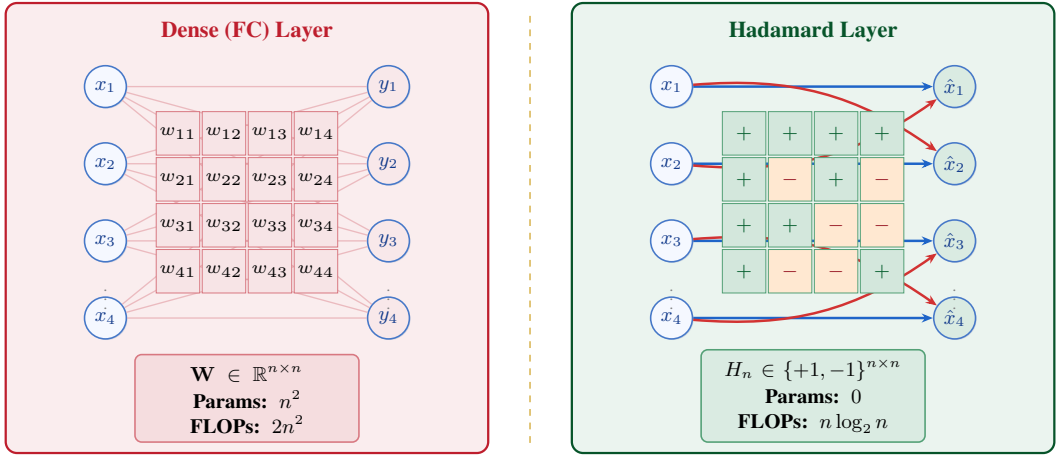

\centering
\includestandalone[width=\textwidth]{dense_hada}
% \caption{Dense vs Hadamard Layer comparison}
\caption{Comparative diagram of dense and Hadamard layers.}
\label{fig:dense_vs_hadamard}
\end{figure}

In this work, we revisit a fundamental yet underexplored component of the attention mechanism: the dense $d \times d$ projection used to combine the outputs of attention heads. While prior efforts have focused on reducing redundancy within attention heads or MLP blocks, the head-mixing projection itself has received comparatively little scrutiny. We challenge the assumption that a full dense projection is necessary for effective head combination, and propose replacing it with a learned Hadamard transformation. This structured alternative substantially reduces both parameter count and computational cost, while achieving performance comparable to the standard dense projection. A schematic comparison between the standard dense projection and the proposed Hadamard-based layer is illustrated in Figure~\ref{fig:dense_vs_hadamard}.
\section{Related Work}

\paragraph{Parameter Efficiency in Attention Mechanisms.}
Early efforts toward reducing attention cost focused on key-value sharing across heads. Shazeer~\cite{shazeer2019fast} introduced \emph{multi-query attention} (MQA), in which all heads share a single key-value projection, yielding substantial memory and latency reductions at the cost of modest quality degradation. Ainslie et al.~\cite{ainslie2023gqa} proposed \emph{grouped-query attention} (GQA), which interpolates between MHA and MQA by partitioning query heads into groups sharing corresponding key-value heads, largely recovering full MHA accuracy while preserving inference efficiency. Together, these works establish that judicious parameter sharing across heads can yield meaningful efficiency gains, though overly aggressive sharing may compromise expressivity.

\paragraph{Redundancy Across Attention Heads.}
Cordonnier et al.~\cite{cordonnier2020collaborate} provided empirical evidence that a significant fraction of attention heads learn redundant key and query projections, motivating \emph{collaborative multi-head attention}, which shares portions of key-query projections across heads with negligible accuracy loss. Jin et al.~\cite{jin2024moh} approached head redundancy from a sparsity perspective via \emph{mixture-of-heads} (MoH) attention, treating heads as experts and dynamically routing each token to a subset of heads, achieving competitive accuracy while activating only 50\%--90\% of available heads.

\paragraph{Structured Parameterizations.}
Wei et al.~\cite{wei2024building} demonstrated that low-rank and block-diagonal constraints on feedforward weight matrices can substantially reduce parameter counts with minimal accuracy degradation, establishing that structured matrix families are effective replacements for dense weights across Transformer components.

\section{Motivation}

\begin{figure}[ht]
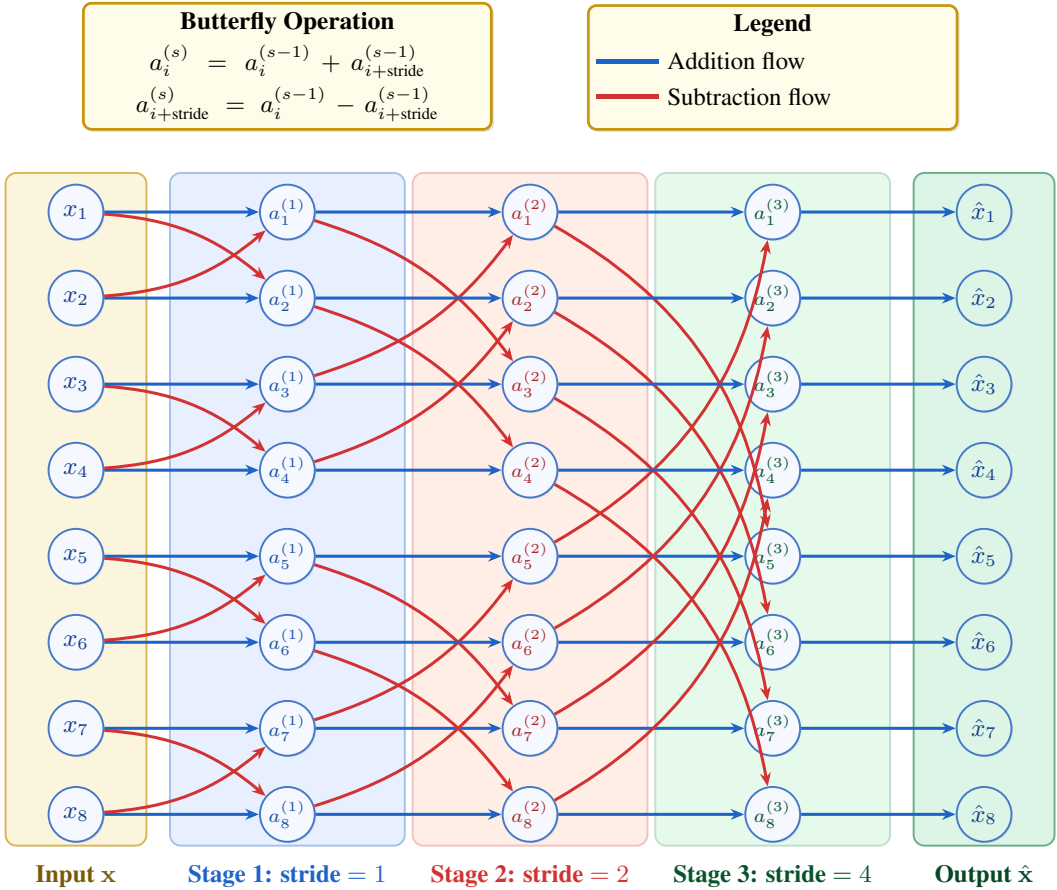

\centering
\includestandalone[width=\textwidth]{butterfly}
\caption{Computational flow of the Fast Walsh-Hadamard Transform (FWHT) showing the butterfly network structure with three stages}
\label{fig:CFlow}
\end{figure}

Multi-head attention (MHA) derives its expressive power from attending to multiple representation subspaces in parallel. Given an input sequence, each attention head independently computes a weighted combination of values, producing a set of head-specific representations. These representations are concatenated and subsequently transformed by a dense linear projection, commonly referred to as the output projection. This projection plays a critical role: it enables interaction across heads, re-projects the concatenated representation back into the model dimension, and ensures compatibility with the residual connection that follows the attention block.

Despite its functional importance, the output projection constitutes a substantial fraction of the parameters within the attention module. As Transformer models scale, this dense transformation increasingly contributes to over-parameterization and computational cost. Empirical evidence from prior work suggests that attention heads exhibit a high degree of redundancy, and that full, unconstrained linear mixing across heads may not be strictly necessary to preserve model performance. This observation motivates the search for alternative head-mixing mechanisms that retain expressive capacity while reducing parameter count and computational overhead.

The Hadamard matrix presents a principled answer to the problem of projection overhead. As a fixed, parameter-free orthogonal transform, the Walsh–Hadamard Transform (WHT) uniformly mixes all input dimensions through a butterfly-structured computation (Figure~\ref{fig:CFlow}), requiring no learned weights while preserving the $\ell_2$ norm of its input. Crucially, this structured mixing is not merely a 
computational convenience; it imposes a specific inductive bias on head interaction.

Because the WHT couples every head output with every other through a fixed, maximally spread orthogonal basis, it encourages the model to distribute information across heads in a globally coherent manner. Rather than permitting arbitrary redundancy through unconstrained linear recombination, this structure implicitly regularizes the attention module: heads are incentivized to learn complementary, non-overlapping representations, as only such representations can be efficiently preserved and distinguished under a fixed orthogonal mixing. Our approach aligns with the rationale for orthogonal initialization in deep networks: structured inductive biases that encourage representational diversity in the absence of explicit supervision.

From a computational perspective, the WHT admits an $\mathcal{O}(d \log d)$ 
butterfly factorization, replacing the $\mathcal{O}(d^2)$ dense matrix multiplication with a cascade of lightweight addition and subtraction operations. Replacing the output projection with this transform eliminates a parameter matrix that typically accounts for approximately 25\% of the total attention parameters in a standard Multi-Head Attention (MHA) block, without introducing new hyperparameters or architectural complexity.

In this work, we propose replacing the dense output projection with a structured transformation based on the Hadamard matrix. By doing so, we preserve the essential role of head interaction and dimensional consistency required for residual connections, while significantly reducing the number of learnable parameters. As illustrated in Figure \ref{fig:CFlow}, (showing a three-stage butterfly computation for $d=8$), this approach challenges the assumption that full dense mixing is necessary for effective head aggregation, suggesting that standard projections are unnecessarily overparameterized.

\section{Methodology}
\subsection{Architecture Overview}

We propose a modified Multi-Head Attention (MHA) block in which the standard dense output projection is replaced by a structured, parameter-free mixing operation based on the Walsh-Hadamard Transform (WHT). All other components, including the query ($W_Q$), key ($W_K$), and value ($W_V$) projections, as well as the per-head attention computation, remain identical to the transformer baseline.

Given an input $X \in \mathbb{R}^{T \times d_{\text{model}}}$, the concatenated outputs of the attention heads are denoted as $Y = \text{Concat}(\text{Attn}_1, \dots, \text{Attn}_h) \in \mathbb{R}^{T \times d_{\text{model}}}$. In a standard transformer, these are mixed via a learned dense projection $W_O \in \mathbb{R}^{d_{\text{model}} \times d_{\text{model}}}$. We instead formulate the output as:

\begin{equation}
\text{MHA}_{\text{Had}}(X) = \boldsymbol{\alpha} \odot (Y \mathcal{H}) + \boldsymbol{\beta}
\end{equation}

where $\mathcal{H} \in \mathbb{R}^{d_{\text{model}} \times d_{\text{model}}}$ is the normalized Walsh-Hadamard matrix satisfying $\mathcal{H}^\top \mathcal{H} = I$, and $\odot$ denotes the Hadamard (element-wise) product. The vectors $\boldsymbol{\alpha}, \boldsymbol{\beta} \in \mathbb{R}^{d_{\text{model}}}$ represent trainable diagonal scaling and bias parameters, respectively. 

% The WHT acts as a fixed orthogonal backbone that ensures global, information-preserving cross-head interaction through a series of $O(d \log d)$ butterfly operations. By replacing the $O(d^2)$ dense projection with this structured transform, we eliminate approximately 25\% of the parameters within the attention block. This design challenges the necessity of unconstrained linear recombination in MHA, suggesting that the inductive bias of a fixed orthogonal mixing, combined with lightweight affine rescaling, is sufficient to maintain expressive capacity while significantly improving computational efficiency.

\subsection{Parameter Efficiency Analysis}

A standard Multi-Head Attention (MHA) block is dominated by four weight matrices: the query ($W_Q$), key ($W_K$), value ($W_V$), and output ($W_O$) projections. In a baseline Transformer, each matrix $\in \mathbb{R}^{d_{\text{model}} \times d_{\text{model}}}$ contributes $d_{\text{model}}^2$ parameters, totaling:
\begin{equation}
4 d_{\text{model}}^2
\end{equation}

In our proposed architecture, the $d_{\text{model}}^2$ parameters of the dense $W_O$ matrix are eliminated. The Walsh-Hadamard Transform is parameter-free, requiring only a learned diagonal scaling vector $\boldsymbol{\alpha} \in \mathbb{R}^{d_{\text{model}}}$. The resulting parameter count for the attention block becomes:
\begin{equation}
3 d_{\text{model}}^2 + d_{\text{model}}
\end{equation}

The relative reduction in attention parameters is calculated as:
\begin{equation}
\frac{4d_{\text{model}}^2 - (3d_{\text{model}}^2 + d_{\text{model}})}{4d_{\text{model}}^2} = \frac{d_{\text{model}}^2 - d_{\text{model}}}{4d_{\text{model}}^2}
\end{equation}

Simplifying the expression yields:
\begin{equation}
\frac{1}{4} - \frac{1}{4d_{\text{model}}} \approx 25\%
\end{equation}

For standard model dimensions (e.g., $d_{\text{model}} \geq 512$), the $1/4d_{\text{model}}$ term is negligible. This modification thus provides a constant \textbf{25\% reduction in attention parameters} per block, significantly lowering the model's memory footprint without reducing the dimensionality of the hidden states.

\subsection{Computational Complexity}

For both standard MHA and the proposed method, the dominant computational cost arises from attention score computation:
\[
\mathcal{O}(T^2 d_{\text{model}}),
\]
which remains unchanged.

The difference lies in the \textit{head-mixing} stage. Standard multi-head attention (MHA) performs head mixing using a dense matrix multiplication with computational complexity
\begin{equation}
\mathcal{O}(T d_{\text{model}}^{2}),
\end{equation}
where $T$ is the sequence length and $d_{\text{model}}$ is the embedding dimension.

In contrast, the proposed method replaces this operation with a Fast Walsh--Hadamard Transform (FWHT)~\cite{le2014fastfoodapproximatekernelexpansions}, yielding a complexity of
\begin{equation}
\mathcal{O}(T d_{\text{model}} \log d_{\text{model}}).
\end{equation}

As illustrated in Figure~\ref{fig:flops_comparison}, the forward FLOPs of dense matrix multiplication scale quadratically with the embedding dimension ($d^2$), whereas the FWHT scales as $d \log_2 d$. The comparison is shown across embedding dimensions commonly used in GPT-2 models. Notably, FWHT introduces no additional trainable parameters, as it requires zero stored weights.

\input{table}

\section{Experiments}
% \subsection{Experimental Setup}
% \label{subsec:experimental_setup}
% \paragraph{Hardware and Implementation.}
% All experiments are conducted on \textbf{8 × NVIDIA H100 (80GB)} GPUs under identical hardware conditions across all model variants to ensure fair comparison. All models are implemented in \texttt{PyTorch}, building upon the NanoGPT codebase~\cite{karpathy_nanogpt_2022} (with rope and swiglu) with modifications confined exclusively to the attention module. No additional architectural changes are introduced unless explicitly stated.

% \paragraph{Training Configuration}
% Training is performed in mixed-precision \textbf{bfloat16 (bf16)} using Distributed Data Parallel (DDP) across all 8 GPUs, with global batch sizes reported in tokens. All models are trained with a fixed context length of 
% nctx=1024. We use the AdamW optimizer with b1=0.9, b2=0.95, e=10{-8}, weight decay of 0.1, and gradient clipping at 1. The learning rate follows a cosine decay schedule with linear warmup; per-model learning rates are reported in Table~\ref{tab:gpt3-models}.

\subsection{Experimental Setup}

\textbf{Hardware and Implementation.} All experiments are conducted on a cluster of $8 \times$ NVIDIA H100 (80GB) GPUs. To ensure a fair comparison, all model variants are evaluated under identical hardware conditions. The models are implemented in PyTorch, extending the NanoGPT framework~\cite{karpathy_nanogpt_2022} with modern architectural enhancements including Rotary Positional Embeddings (RoPE) and SwiGLU activation functions. Modifications are confined exclusively to the attention module; no additional architectural changes are introduced unless explicitly stated.

\textbf{Training Configuration.} Training is performed in mixed-precision \texttt{bfloat16} (bf16) using Distributed Data Parallel (DDP). All models are trained with a fixed context length of $n_{\text{ctx}}=1024$. We utilize the AdamW optimizer with $\beta_1=0.9$, $\beta_2=0.95$, and $\epsilon=10^{-8}$, applying a weight decay of 0.1 and gradient clipping at 1.0. The learning rate follows a cosine decay schedule with a linear warmup phase. Detailed architectural specifications and training hyperparameters for each model variant are summarized in Table~\ref{tab:gpt3-models}.

\definecolor{improvement}{rgb}{0.0, 0.42, 0.10}

\begin{table}[t]
\centering
\caption{Sizes, architectures, and learning hyper-parameters of baseline models.
  $\Delta n_{\text{params}}$ reports the parameter reduction of the Hadamard
  variant relative to Baseline (\improve{green} = reduction).}
\label{tab:gpt3-models}
\setlength{\tabcolsep}{6pt}
\renewcommand{\arraystretch}{1.15}
\begin{adjustbox}{max width=\linewidth}
\begin{tabular}{lcccccccc}
\toprule
Model
  & $n_{\text{params}}^{\text{base}}$
  & $n_{\text{params}}^{\text{hada}}$
  & $\Delta n_{\text{params}}$
  & $n_{\text{layers}}$
  & $d_{\text{model}}$
  & $n_{\text{heads}}$
  & Batch Size
  & Learning Rate \\
\midrule
Tiny  & 124M & 117M & \cellcolor[gray]{0.93}\improve{$7.1\text{M}\ (-5.7\%)$}   & 12 & 768  & 12 & 0.25M & $6.0 \times 10^{-4}$ \\
Small & 354M & 329M & \cellcolor[gray]{0.93}\improve{$25.1\text{M}\ (-7.1\%)$}  & 24 & 1024 & 16 & 0.25M & $3.0 \times 10^{-4}$ \\
Base  & 757M & 701M & \cellcolor[gray]{0.93}\improve{$56.6\text{M}\ (-7.5\%)$}  & 24 & 1536 & 16 & 0.50M & $2.5 \times 10^{-4}$ \\
Large & 1.3B & 1.2B & \cellcolor[gray]{0.93}\improve{$100.6\text{M}\ (-7.7\%)$} & 24 & 2048 & 16 & 1.00M & $2.0 \times 10^{-4}$ \\
\bottomrule

\end{tabular}
\end{adjustbox}
\end{table}

\subsection{Models Compared}
\label{subsec:models_compared}

We evaluate the proposed attention mechanism by comparing it against baseline transformer derived from \emph{NanoGPT}~\cite{karpathy_nanogpt_2022}. The baseline model follows a standard decoder-only Transformer architecture and incorporates \emph{Rotary Positional Embeddings (RoPE)~\cite{su2023roformerenhancedtransformerrotary}} and \emph{SwiGLU}~\cite{shazeer2020gluvariantsimprovetransformer} activation functions. This configuration serves as the reference implementation for all experiments.

% \begin{figure}[H]
% \centering
% \begin{subfigure}{0.48\textwidth}
%     \centering
%     \includegraphics[width=\linewidth]{pdfs/neurips_train_loss1.pdf}
%     \caption{Training loss vs.\ steps.}
% \end{subfigure}
% \hfill
% \begin{subfigure}{0.48\textwidth}
%     \centering
%     \includegraphics[width=\linewidth]{plots/val_loss.png}
%     \caption{Validation loss vs.\ FLOPs.}
% \end{subfigure}

% \caption{\textbf{Training dynamics and scaling behavior of dense vs.\ WHT-augmented models.} (a) While structured variants exhibit a marginal reduction in sample efficiency during early training (left), they demonstrate superior compute utilization as training progresses. (b) WHT-augmented models show a \textbf{steeper scaling trend} of validation loss relative to total training FLOPs (right), achieving a lower loss at equivalent compute budgets across all model scales.}
% \label{fig:scaling_results}
% \end{figure}

\paragraph{Baseline Configurations}
The baselines utilize the standard multi-head attention (MHA) formulation with a dense output projection layer. We evaluate distinct model scales to characterize scaling behavior. All architectural hyperparameters, excluding the projection mechanism, are held constant across variants to isolate the impact of our proposed modification.

\paragraph{WHT-Augmented Models}
Our proposed models are identical to the baselines except for a single controlled modification: the dense output projection is replaced by our Hadamard-based head-mixing mechanism. This "drop-in" replacement strategy ensures that observed improvements in parameter efficiency, computational throughput, and scaling performance are directly attributable to the structured orthogonal mixing.

\begin{figure}[t]
\centering
% Left side: The Table
\begin{minipage}[t]{0.5\textwidth}
    \vspace{0pt} % Forces top alignment
    \captionsetup{type=table}
    \caption{\textbf{Better training FLOPs utilization of the WHT-Augmented Models:} We compare dense Transformers trained according to their optimal scaling law.}
    \label{tab:flops-utilization}
    \vspace{0.5em}
    \centering
    \resizebox{\textwidth}{!}{% Resize table to fit minipage width
        \begin{tabular}{lcccc}
        \toprule
        Model & \#Param & Training FLOPs & PPL \\
        \midrule
        Small & 334M & 1.55e+19 & 21.54  \\
        \textbf{Ours} & \textbf{329M} & 1.55e+19 & \textbf{21.39}  \\
        \midrule
        Base & 757M & 6.61e+19 & 17.41  \\
        \textbf{Ours} & \textbf{701M} & 6.61e+19 & \textbf{17.35}  \\
        \midrule
        Large & 1.3B & 2.06e+20 & 16.17  \\
        \textbf{Ours} & \textbf{1.2B} & 2.06e+20 & \textbf{15.70}  \\
        \bottomrule
        \end{tabular}
    }
\end{minipage}
\hfill
% Right side: The Figure
\begin{minipage}[t]{0.48\textwidth}
    \vspace{0pt} % Forces top alignment
    \centering
    \includegraphics[width=\linewidth]{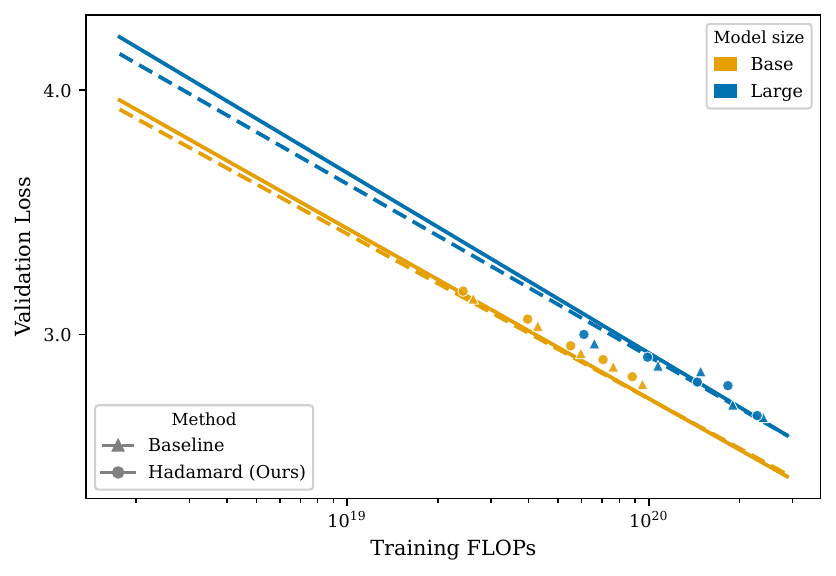}
    \caption{\textbf{Steeper scaling curves with Hadamard projection layers}}
    \label{fig:scaling-curves}
\end{minipage}
\end{figure}

\subsection{Efficiency and Scaling Results}
\label{sec:efficiency_results}

We evaluate the proposed WHT-augmented attention mechanism across two primary axes: (i) \textbf{Predictive Performance} on standard downstream language modeling benchmarks, and (ii) \textbf{Inference Efficiency}, encompassing throughput, latency, and peak memory utilization. All efficiency metrics are measured on a single NVIDIA H100 (80GB) GPU in \texttt{bfloat16} precision.

\subsubsection{Downstream Task Performance}

To ensure that the structural constraints of the Hadamard projection do not degrade model expressivity, we evaluate our method on a suite of zero-shot benchmarks: PIQA~\cite{piqa}, HellaSwag~\cite{hellaswag}, ARC-Easy~\cite{arc}, and BLiMP~\cite{blimp}. 

\begin{figure}[H]
    \centering
    \includegraphics[width=0.85\linewidth]{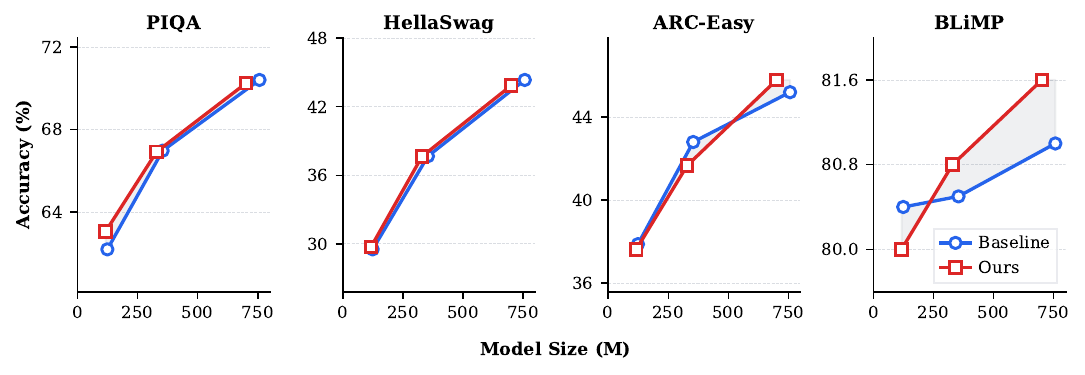}
    \caption{\textbf{Zero-shot performance across model scales.}To ensure a fair comparison between the dense baseline and our structured Hadamard variant, we evaluate performance at a fixed compute budget (isoflops).}
    \label{fig:benchmark_results}
\end{figure}

As shown in Figure~\ref{fig:benchmark_results}, our method achieves comparable accuracy to the baseline across all configurations. This confirms that the orthogonal inductive bias of the Hadamard transform effectively preserves the representational capacity of the attention block while significantly reducing the parameter budget.

\subsubsection{Parameter and Memory Efficiency}

\paragraph{Parameter Scaling.} As shown in the table~\ref{tab:gpt3-models}, the WHT-augmented models yield a consistent reduction in total parameter count, with the relative savings ($\Delta n_{\text{params}}$) scaling monotonically with the model dimension. For the Large variant, our method eliminates over 100M parameters, representing a 7.7\% aggregate reduction while preserving the core Transformer dimensions.

\paragraph{GPU Memory Utilization.} As summarized in Tables~\ref{tab:prefill} and~\ref{tab:decode}, our method consistently reduces peak GPU memory allocation. This reduction in the weight memory footprint directly enables larger batch sizes within the same hardware constraints, which is critical for high-throughput serving workloads.

\subsubsection{Inference Throughput and Latency}

We stress-test the implementation across the two primary stages of LLM inference: the compute-bound \textit{prefill} phase and the memory-bandwidth-bound \textit{decode} phase.

\begin{figure}[t]
    \centering
    \includegraphics[width=1\linewidth]{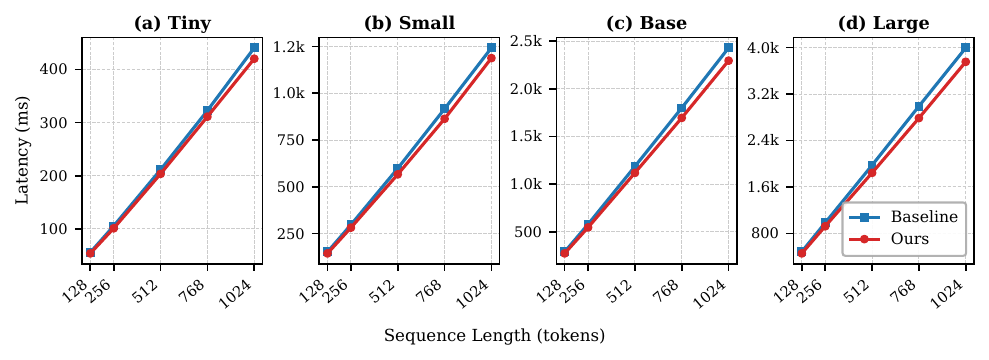}
    \caption{\textbf{Prefill performance analysis(Batch Size = 64).} (a) Our method consistently reduces latency as sequence length increases}
    \label{fig:latency_results}
\end{figure}

\paragraph{Prefill Phase} Table~\ref{tab:prefill} and Figures~\ref{fig:latency_results} report results for the prefill stage. The reduction in arithmetic complexity from $O(d^2)$ to $O(d \log d)$ results in throughput gains of up to 7.0\% for the Large model variant. These gains are most pronounced in large-batch and long-sequence configurations, where the memory-bandwidth-bound regime dominates.

\begin{table}[t]
\centering
\caption{%
  \textbf{Prefill} benchmark: \textbf{Baseline} vs.\ \textbf{Hadamard}
  (batch\,=\,1024, input\,=\,64\,tokens).
  Mean\,$\pm$\,std over 3 runs (50 iters each).
  $\Delta$ rows: Hadamard $-$ Baseline; \improve{green} = improvement
}
\label{tab:prefill}
\setlength{\tabcolsep}{6pt}
\renewcommand{\arraystretch}{1.2}
\begin{adjustbox}{max width=\columnwidth}
\begin{tabular}{@{} ll rr rr r @{}}
\toprule
\textbf{Size} & \textbf{Model}
  & \multicolumn{2}{c}{\textbf{Latency (ms/tok)}}
  & \multicolumn{2}{c}{\textbf{Throughput (tok/s)}}
  & \textbf{Mem (MB)} \\
\cmidrule(lr){3-4}\cmidrule(lr){5-6}
& & mean & $\pm$std & mean & $\pm$std & \\
\midrule
 
\multirow{3}{*}{\textbf{Tiny}}
  & Baseline & 33.157 & 0.029 & 150{,}126 & 29 & 16{,}264 \\
  & Hadamard & 32.366 & 0.011 & 156{,}133 & 13 & 16{,}227 \\
\rowcolor{deltarow}
  & $\Delta$
    & \multicolumn{2}{c}{\improve{$-0.791\ (-2.4\%)$}}
    & \multicolumn{2}{c}{\improve{$+6{,}007\ (+4.0\%)$}}
    & \improve{$-37$} \\[1pt]
 
\multirow{3}{*}{\textbf{Small}}
  & Baseline &  84.699 & 0.031 & 53{,}154 &  5 & 21{,}330 \\
  & Hadamard &  83.204 & 0.019 & 55{,}956 & 14 & 21{,}233 \\
\rowcolor{deltarow}
  & $\Delta$
    & \multicolumn{2}{c}{\improve{$-1.495\ (-1.8\%)$}}
    & \multicolumn{2}{c}{\improve{$+2{,}802\ (+5.3\%)$}}
    & \improve{$-97$} \\[1pt]
 
\multirow{3}{*}{\textbf{Base}}
  & Baseline & 129.595 & 0.074 & 26{,}921 & 13 & 26{,}504 \\
  & Hadamard & 126.442 & 0.052 & 28{,}585 & 15 & 26{,}289 \\
\rowcolor{deltarow}
  & $\Delta$
    & \multicolumn{2}{c}{\improve{$-3.153\ (-2.4\%)$}}
    & \multicolumn{2}{c}{\improve{$+1{,}664\ (+6.2\%)$}}
    & \improve{$-215$} \\[1pt]
 
\multirow{3}{*}{\textbf{Large}}
  & Baseline & 174.158 & 0.055 & 16{,}188 & 8 & 32{,}240 \\
  & Hadamard & 169.501 & 0.047 & 17{,}318 & 11 & 31{,}857 \\
\rowcolor{deltarow}
  & $\Delta$
    & \multicolumn{2}{c}{\improve{$-4.657\ (-2.7\%)$}}
    & \multicolumn{2}{c}{\improve{$+1{,}130\ (+7.0\%)$}}
    & \improve{$-383$} \\
 
\bottomrule
\end{tabular}
\end{adjustbox}
\end{table}

% \begin{figure}[H]
%     \centering
%     \begin{subfigure}{0.48\textwidth}
%         \centering
%         \includegraphics[width=\linewidth]{pdfs/prefill_1.pdf}
%         \caption{Prefill Throughput ($BS=128$)}
%     \end{subfigure}
%     \hfill
%     \begin{subfigure}{0.48\textwidth}
%         \centering
%         \includegraphics[width=\linewidth]{pdfs/prefill_2.pdf}
%         \caption{Speedup Heatmap}
%     \end{subfigure}
%     \caption{\textbf{Prefill performance analysis.} (a) Our method consistently reduces latency as sequence length increases. (b) The speedup grows monotonically with the batch $\times$ sequence length product, highlighting the benefits of structured projections in high-utilization regimes.}
%     \label{fig:prefill_analysis}
% \end{figure}

\begin{figure}[H]
    \centering
    \includegraphics[width=1\linewidth]{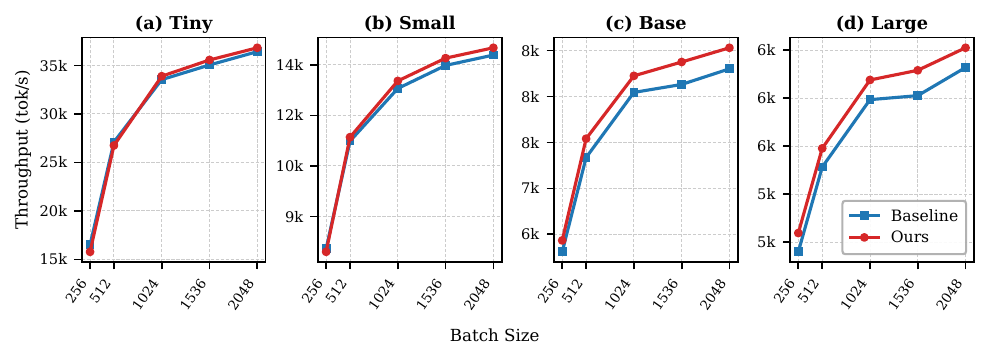}
    \caption{\textbf{Decode dynamics ($T=128$).} Throughput improvements scale with batch size, reflecting the reduced memory pressure of the WHT-based projection during autoregressive generation.}
    \label{fig:decode_throughput}
\end{figure}

\paragraph{Decode Phase.} For autoregressive decoding (Table~\ref{tab:decode}), our method achieves consistent throughput improvements (up to 3.8\%) that scale with batch size. As illustrated in Figure~\ref{fig:decode_throughput}, the structured projection improves memory-bandwidth utilization in the single-token generation regime ($T=1$), which is the primary bottleneck in production environments.

\begin{table}[H]
\centering
\caption{%
  \textbf{Decode} benchmark: \textbf{Baseline} vs.\ \textbf{Hadamard}
  (batch\,=\,2048, gen\,=\,64\,tokens).
  Mean\,$\pm$\,std over 3 runs (50 iters each).
  $\Delta$ rows: Hadamard $-$ Baseline; \improve{green} = improvement
}
\label{tab:decode}
\setlength{\tabcolsep}{6pt}
\renewcommand{\arraystretch}{1.2}
\begin{adjustbox}{max width=\columnwidth}
\begin{tabular}{@{} ll rr rr r @{}}
\toprule
\textbf{Size} & \textbf{Model}
  & \multicolumn{2}{c}{\textbf{Latency (ms/tok)}}
  & \multicolumn{2}{c}{\textbf{Throughput (tok/s)}}
  & \textbf{Mem (MB)} \\
\cmidrule(lr){3-4}\cmidrule(lr){5-6}
& & mean & $\pm$std & mean & $\pm$std & \\
\midrule
 
\multirow{3}{*}{\textbf{Tiny}}
  & Baseline &  39.662 & 0.033 & 52{,}820 & 49 & 7{,}299 \\
  & Hadamard &  39.002 & 0.007 & 53{,}730 &  7 & 7{,}259 \\
\rowcolor{deltarow}
  & $\Delta$
    & \multicolumn{2}{c}{\improve{$-0.660\ (-1.7\%)$}}
    & \multicolumn{2}{c}{\improve{$+910\ (+1.7\%)$}}
    & \improve{$-40$} \\[1pt]
 
\multirow{3}{*}{\textbf{Small}}
  & Baseline & 105.232 & 0.218 & 19{,}978 &  1 & 16{,}625 \\
  & Hadamard & 102.998 & 0.229 & 20{,}409 & 12 & 16{,}529 \\
\rowcolor{deltarow}
  & $\Delta$
    & \multicolumn{2}{c}{\improve{$-2.234\ (-2.1\%)$}}
    & \multicolumn{2}{c}{\improve{$+431\ (+2.2\%)$}}
    & \improve{$-96$} \\[1pt]
 
\multirow{3}{*}{\textbf{Base}}
  & Baseline & 197.373 & 0.070 & 10{,}723 &  9 & 25{,}540 \\
  & Hadamard & 189.879 & 0.241 & 11{,}135 &  2 & 25{,}324 \\
\rowcolor{deltarow}
  & $\Delta$
    & \multicolumn{2}{c}{\improve{$-7.494\ (-3.8\%)$}}
    & \multicolumn{2}{c}{\improve{$+412\ (+3.8\%)$}}
    & \improve{$-216$} \\[1pt]
 
\multirow{3}{*}{\textbf{Large}}
  & Baseline & 262.405 & 0.449 & 8{,}025 & 5 & 35{,}016 \\
  & Hadamard & 252.954 & 0.085 & 8{,}312 & 4 & 34{,}632 \\
\rowcolor{deltarow}
  & $\Delta$
    & \multicolumn{2}{c}{\improve{$-9.451\ (-3.6\%)$}}
    & \multicolumn{2}{c}{\improve{$+287\ (+3.6\%)$}}
    & \improve{$-384$} \\
 
\bottomrule
\end{tabular}
\end{adjustbox}
\end{table}

\section{Discussion and Limitations}

While the Hadamard-based head-mixing layer significantly reduces the theoretical operation count, its practical latency is currently constrained by implementation overhead. Standard General Matrix Multiply (GEMM) kernels benefit from decades of hardware-level optimization and operator fusion, whereas our current Fast Walsh-Hadamard Transform (FWHT) and affine scaling implementation utilizes relatively unoptimized primitives. Consequently, the empirical training speedups do not yet fully reflect the asymptotic $O(d \log d)$ complexity gains. We anticipate that specialized hardware-aware kernels, such as those implemented in Triton or custom CUDA, would allow the practical throughput of our approach to more closely match its theoretical advantages. Furthermore, exploring the interaction between orthogonal mixing and different normalization techniques (e.g., RMSNorm) remains a promising avenue for future work.

\section{Conclusion}

In this work, we proposed an efficient head-mixing mechanism for Multi-Head Attention that replaces the $d \times d$ dense output projection with a parameter-free Hadamard-based transform followed by a lightweight affine layer. Although this modification is confined to a single component of the attention block, its impact on parameter efficiency and memory footprint scales favorably with model width. As Transformer architectures continue to grow in dimension, the cumulative savings provided by structured projections become increasingly vital. Our results suggest that fully dense mixing is not a prerequisite for effective head aggregation, offering a more sustainable path for deploying high-performance models in resource-constrained environments.

\bibliographystyle{plain}
\bibliography{references}

\end{document}

%% file: dense_hada.tex
\begin{tikzpicture}[remember picture]

%% ═══════════════════════════════════════════════════════════
%%  SECTION 3 – Dense Layer vs. Hadamard Layer Comparison
%% ═══════════════════════════════════════════════════════════

\begin{scope}[yshift=-15.4cm]

% ──────────────────────────────────────────────────────────
%  LEFT: Dense Layer diagram
% ──────────────────────────────────────────────────────────
\node[rectangle, rounded corners=5pt,
      fill=DenseRed!8, draw=DenseRed, line width=1pt,
      minimum width=7.5cm, minimum height=7.0cm]
  (densebox) at (-4.4,-3.8) {};
\node[font=\normalsize\bfseries, text=DenseRed]
  at (-4.4,-0.75) {Dense (FC) Layer};

% Input nodes (4 for illustration) - centered
\foreach \i/\yi in {1/0, 2/-1.2, 3/-2.4, 4/-3.6}{
  \node[signode, minimum size=6.5mm] (di\i) at (-6.6, \yi-1.6) {$x_\i$};
}
\node[font=\scriptsize, text=gray] at (-6.6,-4.8) {$\vdots$};

% Output nodes - centered
\foreach \i/\yi in {1/0, 2/-1.2, 3/-2.4, 4/-3.6}{
  \node[signode, minimum size=6.5mm, fill=DenseRed!10]
    (do\i) at (-2.2, \yi-1.6) {$y_\i$};
}
\node[font=\scriptsize, text=gray] at (-2.2,-4.8) {$\vdots$};

% Dense connections (all-to-all)
\foreach \i in {1,...,4}{
  \foreach \j in {1,...,4}{
    \draw[draw=DenseRed!40, line width=0.6pt, opacity=0.7] (di\i) -- (do\j);
  }
}

% Calculate center point between input and output layers
\coordinate (center) at ($(di2)!0.5!(do2)$);
\coordinate (centerbot) at ($(di3)!0.5!(do3)$);
\coordinate (wmatloc) at ($(center)!0.5!(centerbot)$);

% W matrix visual (4x4) - centered in network
\begin{scope}[shift={(wmatloc)}, scale=1.3, transform shape]
\foreach \r in {0,...,3}{
  \foreach \c in {0,...,3}{
    \node[densecell] at (\c*0.55-0.825, -\r*0.55+0.825) {$w_{\pgfmathparse{int(\r+1)}\pgfmathresult\pgfmathparse{int(\c+1)}\pgfmathresult}$};
  }
}
\end{scope}

% Weight matrix annotation - inside box at bottom
\node[rectangle, rounded corners=3pt,
      fill=DenseRed!15, draw=DenseRed!60, line width=0.7pt,
      text width=3.2cm, align=center, inner sep=4pt,
      above=5pt of densebox.south]
  (wmat) {
  \footnotesize
  $\mathbf{W} \in \mathbb{R}^{n \times n}$\\[2pt]
  \textbf{Params:} $n^2$\\
  \textbf{FLOPs:} $2n^2$
};

% ──────────────────────────────────────────────────────────
%  RIGHT: Hadamard Layer diagram
% ──────────────────────────────────────────────────────────
\node[rectangle, rounded corners=5pt,
      fill=HadamardGreen!8, draw=HadamardGreen!70!black, line width=1pt,
      minimum width=7.5cm, minimum height=7.0cm]
  (hadbox) at (4.4,-3.8) {};
\node[font=\normalsize\bfseries, text=HadamardGreen!70!black]
  at (4.4,-0.75) {Hadamard Layer};

% Input nodes - centered
\foreach \i/\yi in {1/0, 2/-1.2, 3/-2.4, 4/-3.6}{
  \node[signode, minimum size=6.5mm] (hi\i) at (2.2, \yi-1.6) {$x_\i$};
}
\node[font=\scriptsize, text=gray] at (2.2,-4.8) {$\vdots$};

% Output nodes - centered
\foreach \i/\yi in {1/0, 2/-1.2, 3/-2.4, 4/-3.6}{
  \node[signode, minimum size=6.5mm, fill=HadamardGreen!15]
    (ho\i) at (6.6, \yi-1.6) {$\hat{x}_\i$};
}
\node[font=\scriptsize, text=gray] at (6.6,-4.8) {$\vdots$};

% FWHT butterfly connections (n=4, 2 stages)
% Stage 1: (1,2),(3,4)
\draw[flowA] (hi1) -- (ho1);
\draw[flowA] (hi2) -- (ho2);
\draw[crossA, bend left=20] (hi1) to (ho2);
\draw[crossA, bend right=20] (hi2) to (ho1);
\draw[flowA] (hi3) -- (ho3);
\draw[flowA] (hi4) -- (ho4);
\draw[crossA, bend left=20] (hi3) to (ho4);
\draw[crossA, bend right=20] (hi4) to (ho3);

% Calculate center point between input and output layers
\coordinate (hcenter) at ($(hi2)!0.5!(ho2)$);
\coordinate (hcenterbot) at ($(hi3)!0.5!(ho3)$);
\coordinate (hmatloc) at ($(hcenter)!0.5!(hcenterbot)$);

% H matrix visual (4x4) – hardcoded ±1 pattern in the center
\begin{scope}[shift={(hmatloc)}, scale=1.3, transform shape]
% Row 0: +1 +1 +1 +1
\node[hcellp] at (-0.825, 0.825) {$+$}; \node[hcellp] at (-0.275, 0.825) {$+$};
\node[hcellp] at ( 0.275, 0.825) {$+$}; \node[hcellp] at ( 0.825, 0.825) {$+$};
% Row 1: +1 -1 +1 -1
\node[hcellp] at (-0.825, 0.275) {$+$}; \node[hcellm] at (-0.275, 0.275) {$-$};
\node[hcellp] at ( 0.275, 0.275) {$+$}; \node[hcellm] at ( 0.825, 0.275) {$-$};
% Row 2: +1 +1 -1 -1
\node[hcellp] at (-0.825,-0.275) {$+$}; \node[hcellp] at (-0.275,-0.275) {$+$};
\node[hcellm] at ( 0.275,-0.275) {$-$}; \node[hcellm] at ( 0.825,-0.275) {$-$};
% Row 3: +1 -1 -1 +1
\node[hcellp] at (-0.825,-0.825) {$+$}; \node[hcellm] at (-0.275,-0.825) {$-$};
\node[hcellm] at ( 0.275,-0.825) {$-$}; \node[hcellp] at ( 0.825,-0.825) {$+$};
% \node[font=\scriptsize, text=HadamardGreen!70!black, above] at (0,1.1)
  % {\footnotesize$H_n$ (fixed $\pm 1$, no storage)};
\end{scope}

% Annotation inside big box but below diagram
\node[rectangle, rounded corners=3pt,
      fill=HadamardGreen!15, draw=HadamardGreen!70, line width=0.7pt,
      text width=3.2cm, align=center, inner sep=4pt,
      above=5pt of hadbox.south]
  (hmat_ann) {
  \footnotesize
  $H_n \in \{+1,-1\}^{n \times n}$\\[2pt]
  \textbf{Params:} $0$\\
  \textbf{FLOPs:} $n\log_2 n$
};

\draw[dashed, AccentGold!60, line width=0.8pt] (0,-0.5) -- (0,-7.2);

\end{scope}

\end{tikzpicture}

%% file: butterfly.tex
\begin{tikzpicture}[remember picture]

%% ═══════════════════════════════════════════════════════════
%%  SECTION 1 – FWHT Butterfly Diagram (n = 8)
%% ═══════════════════════════════════════════════════════════

% ── Stage background panels ───────────────────────────────
\begin{scope}[yshift=-1.4cm]

% Input column bg
\node[stagebg={InputBg}{AccentGold!70}, minimum width=1.8cm, minimum height=8.6cm]
  (bg_in) at (-5.8, -3.8) {};

% Stage 1 bg
\node[stagebg={StageBg1}{ButterflyBlue!40}, minimum width=3.0cm, minimum height=8.6cm]
  (bg_s1) at (-3.1, -3.8) {};

% Stage 2 bg
\node[stagebg={StageBg2}{ButterflyRed!30}, minimum width=3.0cm, minimum height=8.6cm]
  (bg_s2) at (0.0, -3.8) {};

% Stage 3 bg
\node[stagebg={StageBg3}{HadamardGreen!25}, minimum width=3.0cm, minimum height=8.6cm]
  (bg_s3) at (3.1, -3.8) {};

% Output column bg
\node[stagebg={OutputBg}{HadamardGreen!60}, minimum width=1.8cm, minimum height=8.6cm]
  (bg_out) at (5.8, -3.8) {};

% ── Node coordinates ─────────────────────────────────────
% y positions for 8 signals
\foreach \i/\yi in {1/0, 2/-1.1, 3/-2.2, 4/-3.3, 5/-4.4, 6/-5.5, 7/-6.6, 8/-7.7}{
  % Input nodes
  \node[signode] (x\i) at (-5.8, \yi) {$x_{\i}$};
  % Stage-1 output
  \node[signode] (s1\i) at (-3.1, \yi) {};
  % Stage-2 output
  \node[signode] (s2\i) at (0.0, \yi) {};
  % Stage-3 output
  \node[signode] (s3\i) at (3.1, \yi) {};
  % Output nodes
  \node[signode, fill=OutputBg] (y\i) at (5.8, \yi) {$\hat{x}_{\i}$};
}

% ── Stage-1 labels inside nodes ───────────────────────────
\node[font=\scriptsize, text=ButterflyBlue!80!black] at (s11) {$a_1^{(1)}$};
\node[font=\scriptsize, text=ButterflyBlue!80!black] at (s12) {$a_2^{(1)}$};
\node[font=\scriptsize, text=ButterflyBlue!80!black] at (s13) {$a_3^{(1)}$};
\node[font=\scriptsize, text=ButterflyBlue!80!black] at (s14) {$a_4^{(1)}$};
\node[font=\scriptsize, text=ButterflyBlue!80!black] at (s15) {$a_5^{(1)}$};
\node[font=\scriptsize, text=ButterflyBlue!80!black] at (s16) {$a_6^{(1)}$};
\node[font=\scriptsize, text=ButterflyBlue!80!black] at (s17) {$a_7^{(1)}$};
\node[font=\scriptsize, text=ButterflyBlue!80!black] at (s18) {$a_8^{(1)}$};

\node[font=\scriptsize, text=ButterflyRed!80!black] at (s21) {$a_1^{(2)}$};
\node[font=\scriptsize, text=ButterflyRed!80!black] at (s22) {$a_2^{(2)}$};
\node[font=\scriptsize, text=ButterflyRed!80!black] at (s23) {$a_3^{(2)}$};
\node[font=\scriptsize, text=ButterflyRed!80!black] at (s24) {$a_4^{(2)}$};
\node[font=\scriptsize, text=ButterflyRed!80!black] at (s25) {$a_5^{(2)}$};
\node[font=\scriptsize, text=ButterflyRed!80!black] at (s26) {$a_6^{(2)}$};
\node[font=\scriptsize, text=ButterflyRed!80!black] at (s27) {$a_7^{(2)}$};
\node[font=\scriptsize, text=ButterflyRed!80!black] at (s28) {$a_8^{(2)}$};

\node[font=\scriptsize, text=HadamardGreen!70!black] at (s31) {$a_1^{(3)}$};
\node[font=\scriptsize, text=HadamardGreen!70!black] at (s32) {$a_2^{(3)}$};
\node[font=\scriptsize, text=HadamardGreen!70!black] at (s33) {$a_3^{(3)}$};
\node[font=\scriptsize, text=HadamardGreen!70!black] at (s34) {$a_4^{(3)}$};
\node[font=\scriptsize, text=HadamardGreen!70!black] at (s35) {$a_5^{(3)}$};
\node[font=\scriptsize, text=HadamardGreen!70!black] at (s36) {$a_6^{(3)}$};
\node[font=\scriptsize, text=HadamardGreen!70!black] at (s37) {$a_7^{(3)}$};
\node[font=\scriptsize, text=HadamardGreen!70!black] at (s38) {$a_8^{(3)}$};

% ── x → s1 connections (stage 1, stride=1) ───────────────
% Butterfly pairs: (1,2),(3,4),(5,6),(7,8)
\foreach \a/\b in {1/2, 3/4, 5/6, 7/8}{
  % forward (same row)
  \draw[flowA] (x\a) -- (s1\a);
  \draw[flowA] (x\b) -- (s1\b);
  % butterfly cross
  \draw[crossA, bend left=18] (x\a) to (s1\b);
  \draw[crossA, bend right=18] (x\b) to (s1\a);
}

% ── s1 → s2 connections (stage 2, stride=2) ───────────────
% Butterfly pairs: (1,3),(2,4),(5,7),(6,8)
\foreach \a/\b in {1/3, 2/4, 5/7, 6/8}{
  \draw[flowA] (s1\a) -- (s2\a);
  \draw[flowA] (s1\b) -- (s2\b);
  \draw[crossA, bend left=18] (s1\a) to (s2\b);
  \draw[crossA, bend right=18] (s1\b) to (s2\a);
}

% ── s2 → s3 connections (stage 3, stride=4) ───────────────
% Butterfly pairs: (1,5),(2,6),(3,7),(4,8)
\foreach \a/\b in {1/5, 2/6, 3/7, 4/8}{
  \draw[flowA] (s2\a) -- (s3\a);
  \draw[flowA] (s2\b) -- (s3\b);
  \draw[crossA, bend left=25] (s2\a) to (s3\b);
  \draw[crossA, bend right=25] (s2\b) to (s3\a);
}

% ── s3 → output ───────────────────────────────────────────
\foreach \i in {1,...,8}{
  \draw[flowA] (s3\i) -- (y\i);
}

% ── Column labels below ───────────────────────────────────
\node[font=\small\bfseries, text=AccentGold!60!black, below=3pt of bg_in]
  {Input $\mathbf{x}$};
\node[font=\small\bfseries, text=ButterflyBlue, below=3pt of bg_s1]
  {Stage 1: stride $= 1$};
\node[font=\small\bfseries, text=ButterflyRed, below=3pt of bg_s2]
  {Stage 2: stride $= 2$};
\node[font=\small\bfseries, text=HadamardGreen!70!black, below=3pt of bg_s3]
  {Stage 3: stride $= 4$};
\node[font=\small\bfseries, text=HadamardGreen!60!black, below=3pt of bg_out]
  {Output $\hat{\mathbf{x}}$};

% ── Butterfly operation annotation (above columns) ────────
\node[labelbox, text width=5.0cm, align=center,
      fill=yellow!10, draw=AccentGold, line width=1pt,
      drop shadow={shadow xshift=1pt, shadow yshift=-1pt, fill=BoxShadow, opacity=0.4},
      above=15pt of bg_s1.north]
  (bflylabel)
  {\small\textbf{Butterfly Operation}\\[3pt]
   $a_i^{(s)} = a_i^{(s-1)} + a_{i+\text{stride}}^{(s-1)}$\\[2pt]
   $a_{i+\text{stride}}^{(s)} = a_i^{(s-1)} - a_{i+\text{stride}}^{(s-1)}$};

% ── Legend (separate box) ─────────────────────────────────
\node[labelbox, align=center,
      fill=yellow!10, draw=AccentGold, line width=1pt,
      text width=4.5cm, align=left,
      drop shadow={shadow xshift=1pt, shadow yshift=-1pt, fill=BoxShadow, opacity=0.4},
      above=15pt of bg_s3.north] (legend) {
  % \small\textbf{Legend}\\[4pt]
  \small\centering\textbf{Legend}\par\vspace{4pt}
  \begin{tabular}{@{}l@{\hspace{3pt}}l@{}}
  \textcolor{ButterflyBlue}{\rule[0.5ex]{0.8cm}{1.5pt}} & Addition flow\\[3pt]
  \textcolor{ButterflyRed}{\rule[0.5ex]{0.8cm}{1.5pt}} & Subtraction flow\\[3pt]
  % \textcolor{HadamardGreen}{\rule[0.5ex]{1.2cm}{1.5pt}} & Output flow
  \end{tabular}
};

\end{scope}

\end{tikzpicture}

%% file: table.tex
% ── Table ─────────────────────────────────────────────────
% -------------------- TABLE (Conference, Compact, B/W) --------------------
\begin{table}[t]
\centering
\caption{Comparison between standard dense output projections and the proposed Hadamard-based projection. $d$ denotes the model dimension ($d_{\text{model}}$).}
\label{tab:dense_vs_hadamard}
\small
\setlength{\tabcolsep}{8pt}
\renewcommand{\arraystretch}{1.2}
\begin{tabular}{lcc}
\toprule
\textbf{Property} & \textbf{Dense Projection} & \textbf{Hadamard Projection} \\
\midrule
Trainable Parameters & $d^{2}$ & $2d$ \\
Computational Complexity & $\mathcal{O}(d^{2})$ & $\mathcal{O}(d \log d)$ \\
Backward Complexity & $\mathcal{O}(d^{2})$ & $\mathcal{O}(d \log d)$ \\
Weight Memory & $d^{2}$ & $2d$ \\
Theoretical Speedup & $1\times$ & $\mathcal{O}\left(\frac{d}{\log_2 d}\right)\times$ \\
Numerical Stability & Data-dependent & Orthogonal \\
Expressivity & Full-rank map & Fixed orthogonal basis \\
\bottomrule
\end{tabular}
\end{table}

\bigskip

% ── Bar Chart ─────────────────────────────────────────────
\begin{figure}[ht]
\centering
\begin{tikzpicture}
\begin{axis}[
  width=11cm,
  height=7.5cm,
  ybar=0pt,  % No gap between bars
  bar width=18pt,
  symbolic x coords={$d{=}64$, $d{=}256$, $d{=}768$, $d{=}1024$},
  xtick=data,
  x tick label style={font=\normalsize},
  ymin=1,
  ymode=log,
  log basis y=10,
  ylabel={Forward FLOPs (log scale)},
  ylabel style={font=\normalsize},
  ytick={1, 10, 100, 1000, 10000, 100000, 1000000, 10000000, 100000000},
  yticklabels={$10^{0}$,$10^{1}$,$10^{2}$,$10^{3}$,$10^{4}$,
               $10^{5}$,$10^{6}$,$10^{7}$,$10^{8}$},
  yticklabel style={font=\normalsize},
  legend style={
    at={(1.02,0.7)},  % Right side, upper position
    anchor=west,
    font=\normalsize,
    draw=gray!60,
    fill=white,
    inner sep=8pt,
    /tikz/every even column/.append style={column sep=5pt}
  },
  legend cell align=left,
  legend image post style={scale=1.5, yshift=0.1cm},
  grid=both,
  major grid style={line width=0.4pt, draw=gray!40},
  minor grid style={line width=0.25pt, draw=gray!20},
  tick align=outside,
  enlarge x limits=0.20,
  clip=false,
]

% Dense: 2c^2 - c (plotted as c^2 for clarity at scale; difference < 0.1%)
\addplot[fill=DenseRed!70, draw=DenseRed, line width=1pt]
  coordinates {
    ($d{=}64$,   4096)
    ($d{=}256$,  65536)
    ($d{=}768$,  589824)
    ($d{=}1024$, 1048576)
  };
\addlegendentry{\textbf{Dense} $\mathcal{O}(d^{2})$}

% FWHT: c * log2(c)
\addplot[fill=HadamardGreen!65, draw=HadamardGreen!90!black, line width=1pt]
  coordinates {
    ($d{=}64$,    384)
    ($d{=}256$,   2048)
    ($d{=}768$,   7373)
    ($d{=}1024$,  10240)
  };
\addlegendentry{\textbf{FWHT} $\mathcal{O}(d\log_{2} d)$}

% Speedup annotation box - positioned on the right using rel axis cs
\node[
  draw=gray!40,
  fill=white,
  line width=1.2pt,
  rounded corners=3pt,
  inner sep=8pt,
  font=\normalsize,
  align=center,
  anchor=west
] at (rel axis cs:1.02,0.3) {
  \textbf{Speedup:}\\[2pt]
  $\dfrac{768}{\log_2 768} \approx \mathbf{80\times}$
};

\end{axis}
\end{tikzpicture}
\caption{Forward FLOPs for dense matrix multiplication ($d^{2}$) versus the
         Fast Walsh--Hadamard Transform ($d\log_{2}d$) across embedding
         dimensions used in GPT-2 (base: $d{=}768$). FWHT requires zero
         stored weights.}
\label{fig:flops_comparison}
\end{figure}